\documentclass[preprint,12pt,authoryear]{elsarticle}
\usepackage{times}
\usepackage[T1]{fontenc}
\usepackage[utf8]{inputenc}
\usepackage{graphicx}
\usepackage{float}
\usepackage{tikz}
\usetikzlibrary{positioning,arrows.meta,fit,backgrounds,calc, decorations.pathreplacing}
\usepackage{amsmath,amssymb}
\usepackage{booktabs,adjustbox,tabularx}
\usepackage{subcaption}
\usepackage{placeins}

\usepackage{xcolor}

\definecolor{roleA0}{HTML}{5B7A99}    
\definecolor{roleA1}{HTML}{B8923E}    
\definecolor{roleB}{HTML}{A8503C}     
\definecolor{roleC}{HTML}{6E4A6E}     
\definecolor{roleOTHER}{HTML}{8A8A8A} 
\definecolor{trainc}{HTML}{3F6F6A}   

\definecolor{bgDG}{HTML}{F5F2EA}
\definecolor{bgS1}{HTML}{EDF2F5}
\definecolor{bgS2}{HTML}{F0EEF4}

\tikzset{
  unit/.style={draw=#1, fill=#1!8, line width=0.5pt, rounded corners=2pt,
    inner xsep=3pt, inner ysep=4pt, align=center, font=\footnotesize\itshape,
    minimum height=1.0cm, text depth=0pt},
  tag/.style={fill=#1!22, text=#1!75!black, rounded corners=1.5pt,
    font=\scriptsize\bfseries\sffamily, inner xsep=3pt, inner ysep=1.5pt},
  cut/.style={densely dashed, line width=0.5pt, black!55},
}

\usepackage[
    breaklinks,
    colorlinks=true,
    citecolor=blue,
    linkcolor=black,
    urlcolor=blue
]{hyperref}
\usepackage[capitalize]{cleveref}
\journal{}
\begin{document}
\begin{frontmatter}
\title{A domain-general framework for factual-unit extraction and accident-process role classification}
\author[lmbp,lyf]{Aho Yapi}
\ead{aho.yapi@doctorant.uca.fr}
\author[lmbp,iuf]{Pierre Latouche}
\author[lmbp]{Arnaud Guillin}
\author[lyf]{Yan Bailly}
\affiliation[lmbp]{organization={Universit\'e Clermont Auvergne, LMBP, CNRS},city={Aubi\`ere},country={France}}
\affiliation[lyf]{organization={LYF SAS (Cikaba)},city={Clermont-Ferrand},country={France}}
\affiliation[iuf]{organization={Institut Universitaire de France (IUF)},city={Paris},country={France}}

\begin{abstract}
Occupational accident narratives provide detailed descriptions of the
circumstances in which accidents occur and how they unfold. However,
approaches relying on entire narratives or sentences as units of
analysis may combine distinct facts, making it difficult to recover
their individual roles in the accident process. Differences in
reporting practices across industrial domains further complicate this
task. In this paper, we propose a two-stage framework for factual-unit
extraction and accident-process role classification under domain
shift. The first stage adapts Segment Any Text (SaT), a pretrained
sentence-segmentation model, to identify factual-unit boundaries using
supervised learning with low-rank adaptation (LoRA). The second assigns
accident-process roles to the extracted units, comparing fixed
pretrained representations, joint classification training and
supervised representation learning. The framework is evaluated on
17,927 French occupational accident narratives from six domains using
a multi-source leave-one-domain-out protocol. On unseen domains,
adapting SaT substantially improves factual-unit extraction over
pretrained segmentation, in every held-out domain, and this
improvement carries over to the joint recovery of factual units and
their roles. Boundary estimation has a larger effect on the final
representation than the choice of role-classification strategy. The
framework thus provides a basis for analysing recurrent accident
circumstances across reporting domains.
\end{abstract}
\end{frontmatter}

\section{Introduction}
\label{chap3:sec:introduction}

Occupational accident narratives have been used
to classify accidents, identify contributing
factors and analyse recurrent scenarios
\citep{vallmuur2015machine,tixier2016automated,abdat2014extracting}.
These applications differ in the information
they seek to analyse: some concern the accident
as a whole, while others focus on particular
circumstances or events. The appropriate unit
of analysis therefore depends on the information
being sought.

Narrative-level classification assigns categories
to entire accounts, without necessarily locating
the information associated with particular
circumstances or events
\citep{goh2017construction}. Considering
individual sentences provides a finer level
of detail, but a single sentence may still
contain several distinct facts.
For example, the sentence ``While working on a
roof, a worker slipped on a wet surface and
fractured his wrist'' contains several distinct
facts, including the presence of a wet surface
and the injury sustained by the worker.

To distinguish these facts, we consider
\emph{factual units}, defined as contiguous
textual spans, each expressing a distinct
reported fact that can be considered
individually. Several factual units may
therefore occur within the same sentence.
However, identifying individual facts does
not, by itself, indicate their respective
roles in the accident process. In this study,
we distinguish five categories to characterise
these roles: work situations or activities
(A0), adverse conditions explicitly reported
in the narrative (A1), accident events or
deviations (B), consequences (C), and other
reported facts (OTHER). This categorisation
draws on process-oriented accident analysis,
which distinguishes working circumstances,
events and their consequences
\citep{leplat1978accident,monteau1980arbre}.

Obtaining such a representation automatically
involves two related estimation problems:
identifying factual-unit boundaries, which do
not necessarily coincide with sentence boundaries,
and classifying each extracted unit according to
its accident-process role. Since neighbouring
factual units may share the same role while
expressing distinct facts, their boundaries
cannot be determined solely from changes in role.
The two problems are also interdependent: an
incorrect boundary changes the text available
for role classification, while merging distinct
facts into a single interval prevents them from
receiving separate labels.

Occupational accident narratives also vary in
terminology, reporting practices and descriptions
of work activities across sectors and organisations
\citep{khairuddin2022review,pothina2022phrases},
so that models developed on particular domains may
not achieve the same performance on others. We
therefore consider a multi-source domain
generalisation setting, in which models are
trained on several reporting domains and evaluated
on a previously unseen domain
\citep{gulrajani2021domainbed}.

The contributions of this study can be
summarised as follows:

\begin{itemize}

\item We formulate factual-unit extraction and
accident-process role classification as distinct
tasks, accounting for adjacent facts sharing
the same accident-process role.

\item We adapt a pretrained sentence-segmentation
model to factual-unit boundaries and compare
role-classification strategies across unseen
reporting domains.

\item We evaluate both stages under a multi-source
leave-one-domain-out protocol, analysing
cross-role merging errors and their impact
on end-to-end performance.

\end{itemize}

\section{Related work}
\label{chap3:sec:related}

\subsection{Structuring occupational accident narratives}
\label{chap3:subsec:related_accidents}

Occupational accident narratives have been analysed
using structured representations of accident-related
information. For example, \citet{abdat2014extracting}
use expert-defined generic accident factors to
study recurrent scenarios through a Bayesian
network. Since manually identifying such factors
is demanding for large collections of reports,
automatic approaches have been developed to
recover accident-related information.
\citet{bertke2016autocoding} investigate the
automatic assignment of standardised event and
exposure categories to injury narratives, while
other supervised methods classify entire accident
reports according to predefined categories,
including accident types and injury characteristics
\citep{goh2017construction,goldberg2022narratives}.
Beyond narrative-level classification, other
approaches identify specific information within
the text. In construction safety,
\citet{tixier2016automated} use expert-defined
dictionaries and linguistic rules to extract
accident precursors and outcomes, whereas
\citet{Zhang2019ConstructionTextMining}
combine accident-cause classification with
the extraction of objects involved in accidents.
These studies structure accident-related information
at the level of entire narratives or of particular
elements within them, rather than as individual
reported facts.

\subsection{Boundary detection and span extraction}
\label{chap3:subsec:related_spans}

Text segmentation has been studied at different
levels of granularity, from sentences to smaller
discourse units. Sentence boundary detection
can rely on handcrafted linguistic rules to
handle punctuation, abbreviations and other
boundary ambiguities
\citep{sadvilkar2020pysbd}, or on learned
boundary predictions. Segment Any Text (SaT)
uses pretrained representations and supports
adaptation to different domains and segmentation
conventions \citep{frohmann2024segment}.
Discourse segmentation addresses finer textual
units, including elementary discourse units,
whose boundaries follow discourse annotation
conventions
\citep{braud2017discourse,gessler2021discodisco}.
These units need not coincide with individual
reported facts.

In information extraction, textual spans are
identified according to task-specific categories,
such as named entities or event arguments.
Sequence-labelling methods estimate token-level
labels, sometimes using conditional random fields
to model dependencies between successive
predictions \citep{lample2016ner}. Span-based
approaches score candidate intervals using
contextual representations, as in the joint
extraction of entities, relations and events
proposed by \citet{wadden2019dygie}.
Span boundaries and semantic labels can also be
estimated jointly or in successive stages.
Semi-Markov conditional random fields assign
labels to variable-length segments
\citep{sarawagi2004semi}, while
\citet{shen2021locate} and
\citet{pu2022twostage} separate span
localisation from classification in named
entity recognition. In our setting, factual-unit
boundaries are determined by distinct reported
facts, including when neighbouring units
share the same accident-process role.

\subsection{Classification of extracted information}
\label{chap3:subsec:related_roles}

In event argument extraction, semantic roles
are defined in relation to particular events.
\citet{du2020multigranularity} incorporate
different levels of textual context, while
\citet{liu2023rolerelevance} model contextual
clues and dependencies between argument roles.
Similarly, \citet{liang2026event} investigate
span boundaries and argument dependencies
for document-level event extraction. In our
setting, the objective is to assign each
reported fact a role within the accident
process, rather than relative to a particular
event.

The learning strategy also influences how
representations distinguish categories.
Cross-entropy directly optimises class
prediction, whereas metric-learning objectives
organise representations according to their
labels, through relative distance constraints
\citep{hermans2017triplet}, multiple positive
and negative observations within a batch
\citep{khosla2020supcon}, or several learned
centres per class \citep{qian2019softtriple}.
Because factual units sharing an
accident-process role may differ substantially
in semantic content, we compare these
learning strategies for role classification.

\subsection{Generalisation across reporting domains}
\label{chap3:subsec:related_dg}

Domain generalisation addresses domains
unavailable during training. For accident
narratives, \citet{goldberg2022narratives}
investigates the transfer of classification
models to additional datasets, illustrating
the importance of assessing performance beyond
the original training corpus. In text
classification, \citet{tan2022multisourcedg}
investigate supervised contrastive learning
using multiple source domains without access
to target-domain data, and
\citet{gulrajani2021domainbed} highlight the
importance of consistent experimental conditions
and model-selection procedures for reliable
comparisons. For factual-unit extraction and role
classification, changes in reporting
domains may affect both the identification
of factual-unit boundaries and the assignment
of accident-process roles. This motivates
evaluating the two stages separately and
jointly on unseen domains, with model
selection restricted to source-domain data.

\section{Proposed framework}
\label{chap3:sec:framework}

Our objective is to represent an occupational accident
narrative as a collection of individual reported facts,
each associated with a role in the accident process.
We address this objective through two successive
estimation problems: identifying the textual intervals
corresponding to individual units and assigning a role
to each extracted unit. The first problem is treated
through adaptation of a pretrained boundary model;
for the second, we compare alternative strategies for
learning semantic representations under domain shift.
We first define the estimation target, then describe
the two learning stages and their combination.

\subsection{Problem formulation}
\label{chap3:subsec:problem}

Let $x$ denote an accident narrative of length $L$,
measured in characters. Its reference annotation
is represented by
\begin{equation}
\mathcal F(x)=\{(a_k,b_k,r_k)\}_{k=1}^{K},
\qquad 0\leq a_k<b_k\leq L,
\label{chap3:eq:reference_facts}
\end{equation}
where $K$ is the number of annotated factual units,
$[a_k,b_k)$ denotes the character interval of
unit $k$, and $r_k$ its accident-process role.
Each interval corresponds to a contiguous textual
passage expressing a distinct reported fact.
The reference intervals are ordered and
non-overlapping, although they need not cover
every character of the original narrative.

Each role $r_k$ belongs to the set
$\mathcal R=\{\mathrm{A0},\mathrm{A1},\mathrm{B},
\mathrm{C},\mathrm{OTHER}\}$, representing work situations
or activities, adverse conditions, accident events
or deviations, consequences, and other reported
facts, respectively. The category OTHER includes
reported facts not assigned to the four specific
accident-process roles.

The objective is to estimate $\mathcal F(x)$
by identifying factual-unit intervals and
assigning a role to each of them. We denote
by $f_{\theta,\tau}$ the boundary estimator,
where $\theta$ represents the parameters
estimated during adaptation and $\tau$ the
boundary decision threshold. This estimator
produces a collection of intervals
$\widehat{\mathcal U}_{\theta,\tau}(x)$.
A classifier $g_m$ then assigns a role in
$\mathcal R$ to each extracted unit, where
$m$ identifies the classification strategy.

The framework is intended for narratives from reporting contexts,
such as industrial sectors or companies, for which no reference
annotations are available. We therefore formulate the estimation
problem as multi-source domain generalisation. Let $\mathcal D$ denote the set of reporting domains. For a target
domain $q\in\mathcal D$, the source domains are
$\mathcal S_q=\mathcal D\setminus\{q\}$. The parameters $\theta$,
the threshold $\tau$ and the classifier $g_m$ are estimated and
selected using annotated narratives from $\mathcal S_q$ only, and
the resulting estimator of $\mathcal F(x)$ is evaluated on
narratives from $q$. Model selection therefore relies on validation
performance across source domains
(Section~\ref{chap3:subsec:domain_generalisation}).

\subsection{Factual-unit extraction}
\label{chap3:subsec:extraction}

The first stage identifies the textual boundaries
separating individual reported facts. These
boundaries do not necessarily coincide with
sentence endings or changes in accident-process
role. Figure~\ref{chap3:fig:factual_extraction_example}
illustrates this distinction using a real
accident narrative. The narrative contains five
annotated factual units associated with four
roles. The first two units share the role A0
but express distinct facts.

\begin{figure}[!htbp]
\centering
\begin{tikzpicture}[node distance=0.3cm]
  \node[unit=roleA0] (u1) {During the emptying\\of the compactor,};
  \node[unit=roleA0, right=of u1] (u2) {while\\walking,};
  \node[unit=roleA1, right=of u2] (u3) {with the ground\\slippery,};
  \node[unit=roleB,  right=of u3] (u4) {he\\slipped};
  \node[unit=roleC,  right=of u4] (u5) {and twisted his\\left knee.};

  \foreach \u/\r in {u1/A0,u2/A0,u3/A1,u4/B,u5/C}
    \node[tag=role\r, above=2pt of \u.north] (t\u) {\r};

  \foreach \a/\b in {u1/u2,u2/u3,u3/u4,u4/u5}
    \draw[cut] ($(\a.north east)!0.5!(\b.north west)+(0,0.42)$) --
               ($(\a.south east)!0.5!(\b.south west)+(0,-0.12)$);

  \path let \p1=(u1.west), \p2=(u5.east) in
    node[draw=black!35, fill=black!3, rounded corners=2pt, inner sep=5pt,
         text width=\x2-\x1-10pt, align=left, font=\footnotesize,
         anchor=south west] (nar) at ($(u1.north west)+(0,1.55)$)
    {\textsf{\textbf{Accident narrative}}\quad\textit{During the emptying of the compactor,
     while walking, with the ground slippery, he slipped and twisted his left knee.}};

  \draw[-{Stealth[length=5pt]}, line width=0.6pt, black!60]
    (nar.south) -- ++(0,-1.0)
    node[midway, fill=white, inner xsep=4pt, inner ysep=1pt,
         font=\scriptsize\sffamily\bfseries, text=black!85]
    {Segmentation and role assignment};

  \node[font=\scriptsize\sffamily, below=0.4cm of $(u1.south west)!0.5!(u5.south east)$, inner sep=0pt]
    {\textcolor{roleA0}{\rule{6pt}{6pt}}\,A0 Work situation/activity\quad
     \textcolor{roleA1}{\rule{6pt}{6pt}}\,A1 Adverse condition\quad
     \textcolor{roleB}{\rule{6pt}{6pt}}\,B Deviation/event\quad
     \textcolor{roleC}{\rule{6pt}{6pt}}\,C Consequence\quad
     \tikz[baseline=-0.3ex]\draw[cut](0,-0.08)--(0,0.18);\,Boundary};
\end{tikzpicture}
\caption{Reference factual units and accident-process roles in an accident
narrative, translated from French. Dashed lines mark factual-unit boundaries;
the first two units share role A0 but report distinct facts.}
\label{chap3:fig:factual_extraction_example}
\end{figure}

We estimate factual-unit boundaries by adapting Segment Any Text
(SaT), a pretrained model originally developed for sentence
segmentation \citep{frohmann2024segment}. The objective is to identify
boundaries between individual reported facts rather than sentence
endings. For a narrative $x$ of length $L$, SaT estimates boundary
scores at token positions, each associated with the position
immediately after the corresponding token in the original text. Let
$\mathcal J(x)\subseteq\{1,\ldots,L-1\}$ denote the set of candidate
boundary positions. For each $j\in\mathcal J(x)$, let
$Y_j\in\{0,1\}$ indicate whether a factual-unit boundary occurs at
position $j$. The estimated boundary probability is
$p_\theta(j\mid x)=\Pr_\theta(Y_j=1\mid x)$, where $\theta$ denotes
the trainable parameters of the adapted segmentation model.

A candidate position is retained when its estimated probability
exceeds a decision threshold $\tau$. Since these estimates are not
necessarily calibrated, $\tau$ is selected on source-domain validation
data rather than fixed at 0.5. The estimated boundary set, including
the beginning and end of the narrative, is
\begin{equation}
\widehat{\mathcal B}_{\theta,\tau}(x)
=\{0,L\}\cup
\{j\in\mathcal J(x):p_\theta(j\mid x)>\tau\}.
\label{chap3:eq:boundary_set}
\end{equation}
Ordering these boundaries gives $0=c_0<c_1<\cdots<c_M=L$. The
extracted factual units correspond to the intervals between
consecutive boundaries,
\begin{equation}
\widehat{\mathcal U}_{\theta,\tau}(x)
=\bigl\{[c_{i-1},c_i)\bigr\}_{i=1}^{M}.
\label{chap3:eq:extracted_units}
\end{equation}
The number $M$ of extracted units may differ from the number $K$ of
reference units.

To adapt SaT, the reference factual-unit annotations are used to
construct supervised training sequences, tokenised into textual
elements corresponding to words or parts of words. Let
$\widetilde{x}^{(i)}$ denote the $i$-th tokenised training sequence,
containing $T_i$ token positions. For each position
$t\in\{1,\ldots,T_i\}$, the observed indicator $y_{it}\in\{0,1\}$
equals 1 if an annotated factual-unit boundary occurs after that
token and 0 otherwise. It is the observed value of a binary random
variable $Y_{it}$, whose probability
\mbox{$p_{it}=\Pr_\theta(Y_{it}=1\mid\widetilde{x}^{(i)})$} is
estimated by the model. For $n$ training sequences, the model
parameters are estimated by minimising the mean Bernoulli negative
log-likelihood,
\begin{equation}
\mathcal L_{\mathrm{seg}}(\theta)
=-\frac{1}{\sum_{i=1}^{n}T_i}
\sum_{i=1}^{n}\sum_{t=1}^{T_i}
\bigl[\,y_{it}\log p_{it}+(1-y_{it})\log(1-p_{it})\bigr].
\label{chap3:eq:segmentation_loss}
\end{equation}

Rather than updating the entire pretrained encoder, we use low-rank
adaptation (LoRA) \citep{hu2022lora}. LoRA introduces trainable
low-rank updates to selected weight matrices while keeping the
original encoder weights fixed. For a pretrained matrix
$W_0\in\mathbb R^{d_1\times d_2}$, the adapted matrix is written as
\begin{equation}
W=W_0+\frac{\alpha}{k}BA,
\label{chap3:eq:lora_update}
\end{equation}
where $B\in\mathbb R^{d_1\times k}$ and $A\in\mathbb R^{k\times d_2}$
are trainable matrices, $k$ is the adaptation rank and $\alpha$
controls the scale of the update. Since $BA$ has rank at most $k$,
each adapted matrix requires $k(d_1+d_2)$ trainable parameters
instead of $d_1d_2$. The parameters $\theta$ comprise the LoRA
matrices of all adapted weight matrices and the parameters of the
boundary prediction head. They are jointly estimated by minimising
$\mathcal L_{\mathrm{seg}}(\theta)$. Once adapted, SaT estimates
factual-unit boundaries in new narratives, and the extracted
intervals are used as inputs to the accident-process role classifier.

\subsection{Accident-process role classification}
\label{chap3:subsec:role_classification}

The second stage assigns an accident-process role to each extracted
factual unit. Units sharing the same role may describe different
activities, circumstances or events. We compare three strategies for
role classification under domain shift: classification from fixed
pretrained representations, joint training of the encoder and
classifier, and supervised representation learning followed by a
separate classifier. Figure~\ref{chap3:fig:role_strategies}
illustrates these three approaches.

\begin{figure*}[!t]
\centering
\tikzset{
  blk/.style={draw=black!55, fill=white, rounded corners=2pt, align=center,
    text width=20mm, minimum height=9mm, inner sep=2pt, line width=0.5pt,
    font=\scriptsize},
  frozen/.style={blk, fill=black!8, draw=black!45},
  train/.style={blk, fill=trainc!10, draw=trainc},
  io/.style={draw=black!40, fill=black!3, rounded corners=2pt, inner sep=0pt, line width=0.5pt},
  arr/.style={-{Stealth[length=4pt]}, line width=0.5pt, black!60},
  stage/.style={draw=trainc!80, dashed, rounded corners=3pt, line width=0.55pt,
    inner xsep=1.8mm, inner ysep=1.6mm},
  loss/.style={font=\scriptsize\sffamily, text=trainc!80!black, inner sep=1pt},
  rowlab/.style={font=\scriptsize\sffamily\bfseries, text=black!80, anchor=south west, inner sep=0pt},
}
\begin{tikzpicture}[node distance=4mm and 5.5mm]
  \node[frozen] (e1) {Pretrained encoder\\$h_{\phi_0}$, \textit{fixed}};
  \node[frozen, right=of e1, text width=32mm] (r1) {Pretrained representation\\$h_{\phi_0}(u_i)$};
  \node[train,  right=of r1] (c1) {Multinomial\\logistic classifier};
  \node[train, below=12mm of e1] (e2) {Pretrained encoder\\$h_{\phi}$, \textit{fine-tuned}};
  \node[train, right=of e2, text width=32mm] (r2) {Adapted representation\\$h_{\phi}(u_i)$};
  \node[train, right=of r2] (c2) {Multinomial\\logistic classifier};
  \node[train, below=12mm of e2] (e3) {Pretrained encoder\\$h_{\phi}$, \textit{fine-tuned}};
  \node[train] (c3) at (c2 |- e3) {Multinomial\\logistic classifier};
  \node[train, text width=13.5mm, right=2.5mm of e3] (p3) {Projection\\$\Pi_{\psi}$};
  \node[train, text width=19mm, left=4.5mm of c3] (r3) {Projected\\representation $z_i$};

  \node[io, fit={([xshift=-19mm]e1.north west) ([xshift=-6mm]e3.south west)}] (in) {};
  \node[font=\scriptsize, align=center, text width=14mm] at (in.center)
    {\textbf{Factual unit}\\$u_i$\\[3pt]\textit{``with the ground slippery''}};
  \node[io, fit={([xshift=6mm]c1.north east) ([xshift=19mm]c3.south east)}] (out) {};
  \node[font=\scriptsize\bfseries\sffamily, align=center, anchor=south] (ot) at ([yshift=9mm]out.center) {Predicted\\role $\hat r_i$};
  \node[tag=roleA0, below=1.8mm of ot] (ta) {A0};
  \node[tag=roleA1, below=1.4mm of ta] (tb) {A1};
  \node[tag=roleB,  below=1.4mm of tb] (tc) {B};
  \node[tag=roleC,  below=1.4mm of tc] (td) {C};
  \node[tag=roleOTHER, below=1.4mm of td] {OTHER};

  \draw[arr] (e3) -- (p3);
  \draw[arr] (p3) -- (r3);
  \foreach \i in {1,2,3}{
    \draw[arr] (in.east |- e\i.west) -- (e\i.west);
    \ifnum\i<3 \draw[arr] (e\i) -- (r\i);\fi
    \draw[arr] (r\i) -- (c\i);
    \draw[arr] (c\i.east) -- (out.west |- c\i.east);
  }

  \begin{scope}[on background layer]
    \node[stage, fit=(c1)] (s1) {};
    \node[stage, fit=(e2)(r2)(c2)] (s2) {};
    \node[stage, fit=(e3)(p3)(r3)] (s3a) {};
    \node[stage, fit=(c3)(c3 |- e3.north)(c3 |- e3.south)] (s3b) {};
  \end{scope}
  \node[loss, anchor=south east] at (s2.north east)  {$\mathcal{L}_{\mathrm{CE}}(\phi,\omega)$, joint};
  \node[loss, anchor=base west] at ([xshift=1mm,yshift=-4mm]s3a.south west) {Step 1: $\mathcal{L}_{\mathrm{Triplet}}$, $\mathcal{L}_{\mathrm{SupCon}}$ or $\mathcal{L}_{\mathrm{SoftTriple}}$};
  \node[loss, anchor=base east] at ([yshift=-4mm]s3b.south east) {Step 2: classifier fitted on $z_i$};

  \node[rowlab] at ([yshift=2.2mm]e1.north west) {(a) Fixed pretrained representations};
  \node[rowlab] at ([yshift=2.2mm]e2.north west) {(b) Joint training of encoder and classifier};
  \node[rowlab] at ([yshift=2.2mm]e3.north west) {(c) Supervised representation learning, then separate classifier};

  \node[font=\scriptsize\sffamily, anchor=north, inner sep=0pt] at ([yshift=-11mm]r3.south)
    {\tikz\node[frozen, text width=3mm, minimum height=3mm]{};\,Fixed\qquad
     \tikz\node[train, text width=3mm, minimum height=3mm]{};\,Trained\qquad
     \tikz\draw[stage, inner sep=0pt](0,0) rectangle (0.5,0.3);\,Training step};
\end{tikzpicture}
\caption{Three strategies for accident-process role classification.
Grey blocks are fixed and coloured blocks are trained; dashed outlines delimit
training steps and their objectives. In (c), the encoder and projection are first
trained with a supervised representation-learning criterion, then kept fixed while a
separate multinomial logistic classifier is fitted to the projected representations.}
\label{chap3:fig:role_strategies}
\end{figure*}

Let $\mathcal T=\{(u_i,r_i)\}_{i=1}^{N}$ denote the labelled factual
units from the source domains, where $u_i$ is the text of unit $i$,
$r_i\in\mathcal R$ its reference role, and $N$ the number of training
units. A pretrained multilingual encoder $h_\phi$ maps each unit to a
representation in $\mathbb R^d$, where $\phi$ denotes the encoder
parameters and $d$ the representation dimension.

All three strategies assign roles with a multinomial logistic
classifier. In the first two, it is applied to $h_\phi(u_i)$, with
parameters $\omega=\{(\beta_r,\eta_r):r\in\mathcal R\}$, where
$\beta_r\in\mathbb R^d$ and $\eta_r\in\mathbb R$ are the coefficient
vector and intercept associated with role $r$. Denoting by $R_i$ the
random variable representing the role of unit $i$, the estimated
probability \mbox{$p_{\phi,\omega}(r\mid u_i)=\Pr_{\phi,\omega}(R_i=r\mid u_i)$}
is obtained by applying a softmax to the linear scores
$\beta_r^\top h_\phi(u_i)+\eta_r$.

In the first strategy, the encoder remains fixed at its pretrained
parameters $\phi_0$, and only the logistic classifier is estimated.
This provides a reference for assessing the role information
available in pretrained representations without task-specific
adaptation. In the second strategy, the encoder and classifier are
trained jointly by minimising the weighted multiclass cross-entropy
\begin{equation}
\mathcal L_{\mathrm{CE}}(\phi,\omega)
=-\frac{1}{N}\sum_{i=1}^{N}w_{r_i}
\log p_{\phi,\omega}(r_i\mid u_i),
\label{chap3:eq:ce_loss}
\end{equation}
where $w_r>0$ is the weight assigned to role $r$ to account for
unequal class frequencies. This strategy adapts the encoder directly
to role prediction.

In the third strategy, the encoder is adapted using one of three
supervised representation-learning criteria: batch-hard Triplet,
supervised contrastive learning (SupCon), or SoftTriple. The encoder
is combined with a trainable projection
$\Pi_\psi:\mathbb R^d\to\mathbb R^{d_p}$, where $\psi$ denotes its
parameters and $d_p$ the projection dimension. The normalised
projected representation of unit $i$ is denoted by
$z_i\in\mathbb R^{d_p}$. After learning these representations, the
encoder and projection are kept fixed and a separate multinomial
logistic classifier is fitted to $z_i$.

\subsubsection{Batch-hard Triplet}
\label{chap3:subsubsec:triplet}

Batch-hard Triplet learning encourages factual units with the same
role to have more similar representations than units assigned
different roles \citep{hermans2017triplet}. Let
$\mathcal B\subseteq\{1,\ldots,N\}$ denote a mini-batch of training
indices. For each unit $i\in\mathcal B$, called the anchor, let
$\mathcal P_i$ denote the other indices in the mini-batch sharing
role $r_i$, and $\mathcal N_i$ those assigned a different role.

Since the representations $z_i$ have unit norm, we use the cosine
distance $d_{ij}=1-z_i^\top z_j$. The batch-hard formulation selects
the most distant positive and the closest negative for each anchor:
\[
d_i^+=\max_{p\in\mathcal P_i}d_{ip},
\qquad
d_i^-=\min_{a\in\mathcal N_i}d_{ia}.
\]
The soft-margin Triplet criterion is
\begin{equation}
\mathcal L_{\mathrm{Triplet}}(\phi,\psi)
=\frac{1}{|\mathcal B|}\sum_{i\in\mathcal B}
\log\bigl(1+\exp(d_i^+-d_i^-)\bigr).
\label{chap3:eq:triplet_loss}
\end{equation}
Mini-batches are constructed to contain positive and negative
observations for each anchor.

\subsubsection{Supervised contrastive learning}
\label{chap3:subsubsec:supcon}

Supervised contrastive learning uses all same-role observations in
the mini-batch as positives and contrasts each anchor with all other
units, rather than selecting only the hardest positive and negative
\citep{khosla2020supcon}. We use the same mini-batch $\mathcal B$ and
positive sets $\mathcal P_i$ as above. Let
$\mathcal A_i=\mathcal B\setminus\{i\}$ and
$s_{ia}=z_i^\top z_a/T_c$, where $T_c>0$ is a temperature parameter.
The supervised contrastive criterion is
\begin{equation}
\mathcal L_{\mathrm{SupCon}}(\phi,\psi)
=-\frac{1}{|\mathcal B|}\sum_{i\in\mathcal B}
\frac{1}{|\mathcal P_i|}\sum_{p\in\mathcal P_i}
\log\frac{\exp(s_{ip})}
{\sum_{a\in\mathcal A_i}\exp(s_{ia})}.
\label{chap3:eq:supcon_loss}
\end{equation}

\subsubsection{SoftTriple}
\label{chap3:subsubsec:softtriple}

SoftTriple differs from the two preceding criteria by representing
each role with several learned centres rather than relying only on
comparisons between training units \citep{qian2019softtriple}. This
is intended for cases where units assigned the same accident-process
role describe different types of situations or events.

For each role $r\in\mathcal R$, let
$\mu_{r1},\ldots,\mu_{rQ}\in\mathbb R^{d_p}$ denote its $Q$ learned
centres. The similarity between unit $i$ and centre $\ell$ of role
$r$ is $s_{ir\ell}=\cos(z_i,\mu_{r\ell})$. These similarities are
combined to obtain a score for each role,
\[
S_{ir}=\sum_{\ell=1}^{Q}\nu_{ir\ell}\,s_{ir\ell},
\qquad
\nu_{ir\ell}=\frac{\exp(s_{ir\ell}/\gamma)}
{\sum_{m=1}^{Q}\exp(s_{irm}/\gamma)},
\]
where $\gamma>0$ determines how strongly the more similar centres
influence the role score.

To distinguish the reference role from the other roles, SoftTriple
subtracts a margin $\delta\geq0$ from its score during training.
The adjusted score for role $r$ is
\[
v_{ir}^{\mathrm{ST}}
=\lambda\bigl(S_{ir}-\delta\mathbf 1_{\{r=r_i\}}\bigr),
\]
where $\lambda>0$ is a scaling parameter. The resulting multiclass
negative log-likelihood is
\begin{equation}
\mathcal L_{\mathrm{ST}}(\phi,\psi,\mu)
=\frac{1}{N}\sum_{i=1}^{N}
\left[\log\sum_{r\in\mathcal R}
\exp(v_{ir}^{\mathrm{ST}})-v_{ir_i}^{\mathrm{ST}}\right],
\label{chap3:eq:softtriple_loss}
\end{equation}
where $\mu$ denotes the collection of learned centres. The model
also penalises distances between centres associated with the same
role. The complete SoftTriple objective is
\begin{equation}
\mathcal L_{\mathrm{SoftTriple}}(\phi,\psi,\mu)
=\mathcal L_{\mathrm{ST}}(\phi,\psi,\mu)
+\zeta\,\mathcal R_{\mathrm{merge}}(\mu),
\label{chap3:eq:softtriple_regularised}
\end{equation}
where $\zeta\geq0$ controls the strength of this penalty. The
expression of $\mathcal R_{\mathrm{merge}}$ and the centre
normalisation are given in \ref{app:softtriple_details}.

The SoftTriple centres are used to learn representations but do not
enter the final logistic classifier. The comparison therefore
concerns complete learning strategies rather than individual loss
functions. Experimental settings are detailed in
Section~\ref{chap3:subsec:implementation}.

\subsection{Construction of the structured representation}
\label{chap3:subsec:inference}

The two estimation stages are combined to represent
an accident narrative as a collection of textual
intervals with their predicted accident-process roles.
For an interval $[a,b)$, we write $x_{[a,b)}$ for
the corresponding portion of narrative $x$.
Using the estimated boundaries
$0=c_0<c_1<\cdots<c_M=L$, let
$\widehat u_i=x_{[c_{i-1},c_i)}$ denote the text of
extracted unit $i$, and let
$\widehat r_i=g_m(\widehat u_i)$ denote its
predicted role. The resulting structured
representation is

\begin{equation}
\widehat{\mathcal F}_{\theta,\tau,m}(x)
=\bigl\{(c_{i-1},c_i,\widehat r_i)\bigr\}_{i=1}^{M}.
\label{chap3:eq:complete_output}
\end{equation}

Each element records the location of an extracted
unit and the role assigned to it.
Figure~\ref{chap3:fig:framework} summarises
the two-stage estimation procedure.

\begin{figure*}[!t]
\centering
\providecommand{\figL}{\fontsize{8}{9.6}\selectfont}
\providecommand{\figS}{\fontsize{7}{8.4}\selectfont}
\tikzset{
  unit/.style={draw=#1, fill=#1!8, line width=0.5pt, rounded corners=2pt,
    inner xsep=3pt, inner ysep=4pt, align=center, font=\figL\itshape,
    minimum height=1.0cm, text depth=0pt},
  tag/.style={fill=#1!22, text=#1!75!black, rounded corners=1.5pt,
    font=\figS\bfseries\sffamily, inner xsep=3pt, inner ysep=1.5pt},
  panel/.style={draw=black!30, fill=black!2, rounded corners=4pt, line width=0.5pt},
  ptitle/.style={font=\figL\sffamily\bfseries, text=black!85, anchor=north west, inner sep=0pt},
  dom/.style={draw=black!40, fill=white, rounded corners=2pt, align=center,
    text width=18mm, minimum height=8mm, inner sep=1pt, font=\figS\sffamily, line width=0.5pt},
  target/.style={dom, draw=roleB!80!black, line width=0.9pt, fill=roleB!6},
  fixedb/.style={draw=black!50, fill=black!12, rounded corners=2pt, align=center,
    inner sep=2pt, font=\figS, line width=0.5pt},
  trainb/.style={draw=trainc, fill=trainc!14, rounded corners=2pt, align=center,
    inner sep=2pt, font=\figS, line width=0.5pt},
  arr/.style={-{Stealth[length=4.5pt]}, line width=0.6pt, black!60},
  note/.style={font=\figS, text=black!80, inner sep=1pt},
  obj/.style={font=\figS, text=trainc!75!black, inner sep=1pt},
}
\resizebox{\textwidth}{!}{%
\begin{tikzpicture}

\node[panel, fill=bgDG, draw=bgDG!70!black!35, minimum width=12.9cm, minimum height=2.3cm, anchor=north west] (P0) at (0,0) {};
\node[ptitle] at (0.2,-0.15) {Multi-source domain generalisation};

\node[target] (d1) at (1.20,-0.95) {Construction};
\node[dom]    (d2) at (3.30,-0.95) {Metallurgy};
\node[dom]    (d3) at (5.40,-0.95) {Chemicals and\\plastics};
\node[dom]    (d4) at (7.50,-0.95) {Wood and\\textiles};
\node[dom]    (d5) at (9.60,-0.95) {Temporary\\employment};
\node[dom]    (d6) at (11.70,-0.95) {Company\\corpus};

\draw[decorate, decoration={brace, mirror, amplitude=4pt}, black!55, line width=0.5pt]
  ([yshift=-1.5mm]d2.south west) -- ([yshift=-1.5mm]d6.south east)
  node[midway, below=5pt, note, align=center]
  {\textbf{Source domains} $\mathcal S_q$: estimation and model selection};
\node[note, anchor=north, text=roleB!75!black, align=center] at ([yshift=-1.5mm]d1.south)
  {\textbf{Target domain} $q$\\unseen};

\node[panel, fill=white, anchor=north west, minimum width=2.2cm, minimum height=4.15cm] (N) at (0,-3.1) {};
\node[ptitle] at (0.15,-3.25) {Narrative $x$};
\node[font=\figS\itshape, text width=1.9cm, align=flush left, anchor=north west, inner sep=0pt]
  at (0.15,-3.75) {During the emptying of the compactor,
  while walking, with the ground slippery, he slipped and twisted his left knee.};

\node[panel, fill=bgS1, draw=bgS1!60!black!40, anchor=north west, minimum width=5.15cm, minimum height=4.15cm] (S1) at (2.6,-3.1) {};
\node[ptitle] at (2.75,-3.25) {Stage 1: factual-unit extraction $f_{\theta,\tau}$};

\node[fixedb, text width=16.5mm, minimum height=8mm] (sat)  at (3.58,-4.15) {SaT encoder\\\textit{fixed}};
\node[trainb, text width=16.5mm, minimum height=6mm] (lora) at (3.58,-5.0) {LoRA $\tfrac{\alpha}{k}BA$};
\node[trainb, text width=16.5mm, minimum height=6mm] (head) at (3.58,-5.75) {Boundary head};
\draw[black!50, line width=0.5pt] (sat) -- (lora) (lora) -- (head);

\begin{scope}[shift={(4.6,-5.85)}]
  \def\bw{0.085}
  \foreach \h [count=\k] in {.08,.05,.12,.06,.09,.92,.15,.88,.07,.10,.06,.83,.18,.90,.10,.06,.08,.05,.07,.30}{
    \pgfmathsetmacro{\xx}{(\k-1)*0.13}
    \pgfmathparse{\h>0.5 ? 1 : 0}
    \ifnum\pgfmathresult=1
      \fill[black!75] (\xx,0) rectangle ++(\bw,\h*1.0);
    \else
      \fill[black!25] (\xx,0) rectangle ++(\bw,\h*1.0);
    \fi
  }
  \draw[black!50, line width=0.4pt] (-0.05,0) -- (2.6,0);
  \draw[densely dashed, roleB!80!black, line width=0.6pt] (-0.05,0.5) -- (2.6,0.5)
     node[right=1pt, font=\figS, text=roleB!80!black, inner sep=0pt] {$\tau$};
  \node[note, anchor=south west] at (-0.05,0.98) {$p_\theta(j\mid x)$};
  \node[note, anchor=north] at (1.28,-0.02) {$j\in\mathcal J(x)$};
\end{scope}
\draw[arr] (head.east) -- ([xshift=-1mm]4.6,0 |- head.east);

\node[obj, anchor=south] at (3.58,-7.12) {$\mathcal L_{\mathrm{seg}}(\theta)$};
\node[note, anchor=south] at (5.9,-7.12) {$\widehat{\mathcal U}_{\theta,\tau}(x)$};

\node[panel, fill=bgS2, draw=bgS2!60!black!40, anchor=north west, minimum width=4.75cm, minimum height=4.15cm] (S2) at (8.15,-3.1) {};
\node[ptitle] at (8.3,-3.25) {Stage 2: role classification $g_m$};

\node[font=\figS] (ui) at (8.33,-4.55) {$\hat u_i$};
\node[draw=black!45, fill=white, rounded corners=2pt, align=center, inner sep=2pt, font=\figS, line width=0.5pt, minimum width=11mm, minimum height=9mm] (enc) at (9.32,-4.55) {Encoder};
\draw[arr] (ui) -- (enc);

\begin{scope}[shift={(10.05,-5.35)}, scale=0.92]
  \draw[black!35, line width=0.4pt, fill=white, rounded corners=1.5pt] (0,0) rectangle (1.74,1.74);
  \pgfmathsetseed{7}
  \foreach \cx/\cy/\r in {0.42/1.32/A0, 1.32/1.34/A1, 0.40/0.40/B, 1.32/0.42/C}{
    \foreach \k in {1,...,9}{
      \fill[role\r] (\cx+0.17*rand, \cy+0.17*rand) circle (0.032);
    }
  }
  \foreach \k in {1,...,6}{
    \fill[roleOTHER] (0.87+0.42*rand, 0.87+0.42*rand) circle (0.032);
  }
  \draw[densely dashed, black!60, line width=0.5pt] (0.86,0.05) -- (0.90,1.69);
  \draw[densely dashed, black!60, line width=0.5pt] (0.05,0.90) -- (1.69,0.84);
\end{scope}
\draw[arr] (enc) -- (10.03,-4.55);
\node[note, anchor=north] at (10.85,-5.37) {logistic classifier};

\node[tag=roleA0]    (ta) at (12.32,-3.90) {A0};
\node[tag=roleA1]    (tb) at (12.32,-4.225) {A1};
\node[tag=roleB]     (tc) at (12.32,-4.55) {B};
\node[tag=roleC]     (td) at (12.32,-4.875) {C};
\node[tag=roleOTHER] (te) at (12.32,-5.20) {OTHER};
\draw[arr] (11.68,-4.55) -- (11.9,-4.55);

\node[draw=black!25, fill=white, rounded corners=2pt, anchor=north west,
      minimum width=4.45cm, minimum height=1.32cm, inner sep=0pt] (strat) at (8.3,-5.75) {};
\draw[densely dotted, black!50, line width=0.5pt] (enc.south) -- (enc.south |- strat.north);
\node[font=\figS\sffamily\bfseries, text=black!75, anchor=north west, inner sep=0pt]
  at (8.4,-5.84) {Strategy $m$ (encoder training)};
\tikzset{chip/.style={rounded corners=1.5pt, inner xsep=2.5pt, inner ysep=1.5pt, font=\figS, line width=0.5pt, anchor=west}}
\node[chip, draw=black!50, fill=black!12]   (ma) at (8.4,-6.4) {(a) fixed};
\node[chip, draw=trainc,   fill=trainc!14, right=1.2mm of ma] (mb) {(b) joint $\mathcal L_{\mathrm{CE}}(\phi,\omega)$};
\node[chip, draw=trainc,   fill=trainc!14] (mc) at (8.4,-6.8)
  {(c) $\mathcal L_{\mathrm{Triplet}}$, $\mathcal L_{\mathrm{SupCon}}$ or $\mathcal L_{\mathrm{SoftTriple}}$};

\draw[arr] (N.east |- 0,-4.95) -- (S1.west |- 0,-4.95);
\draw[arr] (S1.east |- 0,-4.55) -- (S2.west |- 0,-4.55);
\draw[arr, roleB!70!black] (1.2,-2.12) -- (1.2,-3.1);
\draw[arr] (5.0,-2.3) -- (5.0,-3.1) node[midway, right=1pt, note] {$\theta,\tau$};
\draw[arr] (9.6,-2.3) -- (9.6,-3.1) node[midway, right=1pt, note] {$g_m$};

\node[panel, fill=white, anchor=north west, minimum width=12.9cm, minimum height=2.45cm] (O) at (0,-7.5) {};
\node[ptitle] at (0.2,-7.65) {Structured representation \textmd{$\widehat{\mathcal F}_{\theta,\tau,m}(x)$}};
\draw[arr] (S2.south) -- (S2.south |- O.north);

\begin{scope}[node distance=0.3cm]
  \node[unit=roleA0, anchor=north west] (u1) at (1.0,-8.5) {During the emptying\\of the compactor,};
  \node[unit=roleA0, right=of u1] (u2) {while\\walking,};
  \node[unit=roleA1, right=of u2] (u3) {with the ground\\slippery,};
  \node[unit=roleB,  right=of u3] (u4) {he\\slipped};
  \node[unit=roleC,  right=of u4] (u5) {and twisted his\\left knee.};
  \foreach \u/\r in {u1/A0,u2/A0,u3/A1,u4/B,u5/C}
    \node[tag=role\r, above=2pt of \u.north] {\r};
  \foreach \a/\b in {u1/u2,u2/u3,u3/u4,u4/u5}
    \draw[cut] ($(\a.north east)!0.5!(\b.north west)+(0,0.42)$) --
               ($(\a.south east)!0.5!(\b.south west)+(0,-0.12)$);
\end{scope}

\end{tikzpicture}}
\caption{Overview of the framework, illustrated with Construction as the target
domain. Stage~1 estimates factual-unit boundaries (grey: fixed pretrained weights;
green: trained components) and Stage~2 maps each extracted unit to a representation
and assigns its accident-process role with a multinomial logistic classifier; the
strategies $m$ differ in how the representation is learned. Both stages are
estimated and selected on the source domains and applied to the unseen target
domain. The output shown is illustrative.}
\label{chap3:fig:framework}
\end{figure*}

\section{Experimental design}
\label{chap3:sec:experiments}

We evaluate factual-unit extraction and accident-process
role classification, both separately and jointly, across
six occupational accident reporting domains. The
experimental design follows a multi-source
leave-one-domain-out protocol to assess performance
on unseen domains.

\subsection{Datasets and factual-unit annotations}
\label{chap3:subsec:data}

\begin{figure*}[!t]
\centering
\includegraphics[width=1\textwidth]
{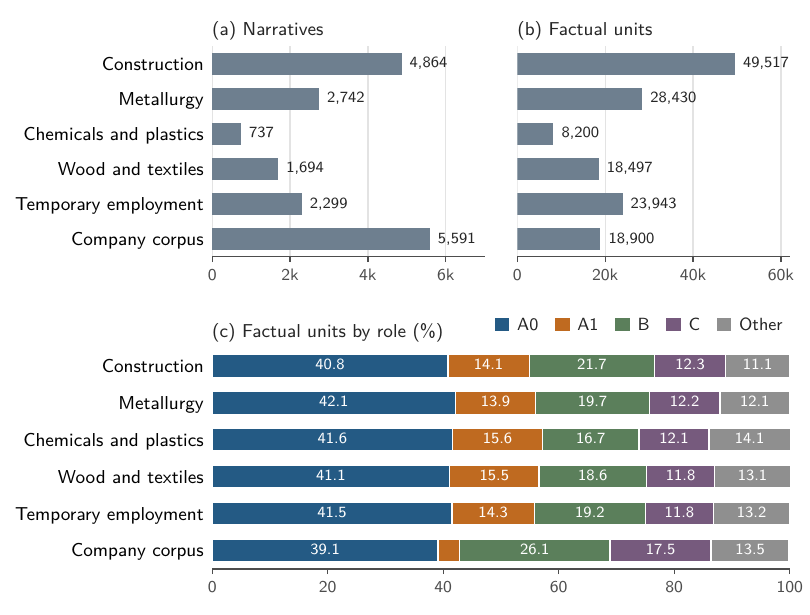}
\caption{Corpus composition by reporting domain: (a) number of narratives, (b) number of annotated factual units, and (c) distribution of factual units across accident-process roles (\%).}
\label{chap3:fig:domain_data}
\end{figure*}

The study comprises 17,927 French occupational accident
narratives from six reporting domains, containing 147,487
annotated factual units. Five sector-specific corpora
are drawn from the EPICEA database, while the sixth
is a separate company-specific corpus. Each unit is
identified by its position in the narrative and assigned
one of the five accident-process roles defined in
Section~\ref{chap3:subsec:problem}. Two experts in
occupational accident analysis produced the reference
annotations. The annotation procedure and the treatment
of ambiguous cases are described in
\ref{chap3:app:data_splits}. Figure~\ref{chap3:fig:domain_data} summarises the corpus sizes
and role distributions. Differences in corpus size
and role frequencies motivate the use of unweighted
domain means and balanced accuracy, respectively.
Additional statistics on narrative and factual-unit
lengths are provided in
Table~\ref{chap3:tab:corpus_lengths}.

\subsection{Multi-source domain generalisation}
\label{chap3:subsec:domain_generalisation}

In each of the six experiments, one domain is held
out for testing, while the other five provide
the training and validation data. Each domain
therefore serves once as the target domain.

Within each source domain, approximately 80\%
of narratives are used for training and 20\%
for validation. The partition is defined at
the narrative level so that factual units
from the same narrative cannot appear in both
subsets. Training data from the five source
domains are pooled, whereas validation
performance is calculated separately for
each domain.

Model selection is based on the average
validation performance across source domains.
For a held-out domain $q$ with source domains
$\mathcal S_q$ (Section~\ref{chap3:subsec:problem}),
let $\kappa\in\{\mathrm{unit},\mathrm{role}\}$
identify the estimation task and
$\mathcal C_{q,\kappa}$ the set of candidate
model configurations. For a configuration
$c\in\mathcal C_{q,\kappa}$, denote by
$V_{q',\kappa}(c)$ its validation score in
source domain $q'$. The selected configuration
is
\begin{equation}
\widehat c_{q,\kappa}
=
\underset{c\in\mathcal C_{q,\kappa}}{\arg\max}
\frac{1}{|\mathcal S_q|}
\sum_{q'\in\mathcal S_q} V_{q',\kappa}(c).
\label{chap3:eq:source_model_selection}
\end{equation}

For segmentation, the model configuration and the
decision threshold are selected using exact-span F1 on the
source-domain validation sets, after removing
surrounding whitespace and punctuation from
predicted and reference intervals. This criterion
requires both boundaries of a reference unit to
be recovered. Test performance is primarily
assessed with factual-unit F1 at an intersection-over-union
threshold of 0.5 (Section~\ref{chap3:subsec:metrics}), which
tolerates limited boundary discrepancies while requiring
substantial overlap. Selection thus emphasises
boundary accuracy, whereas the main test criterion
measures unit recovery under a less restrictive
condition. For role classification, the validation
criterion is balanced accuracy. Averaging scores
by domain prevents larger corpora from dominating
selection. The held-out target domain is not used
for parameter estimation or model selection.

\subsection{Compared learning strategies}
\label{chap3:subsec:implementation}

For factual-unit extraction, we compare four
configurations. PySBD provides a rule-based
sentence-segmentation baseline. Pretrained SaT
is evaluated using a fixed threshold of 0.25, the
default value for \texttt{sat-12l-sm}, and a threshold
calibrated on source-domain validation data. The fourth configuration adapts
SaT to factual-unit boundaries using LoRA.
These comparisons distinguish improvements
obtained through threshold calibration from
those resulting from model adaptation.
For adapted SaT, the model and decision threshold
are selected jointly using source-domain
validation data.

For accident-process role classification, we
compare strategies that differ in how semantic
representations are obtained. Term
frequency--inverse document frequency (TF--IDF)
with multinomial logistic regression (LR)
provides a lexical baseline. The neural
strategies use the same pretrained multilingual
encoder, \textit{Qwen3-Embedding-0.6B}
\citep{zhang2025qwen3embedding}, which produces
1,024-dimensional representations.

We first consider a fixed pretrained encoder
followed by LR. We then consider the adapted
strategies of Section~\ref{chap3:subsec:role_classification}:
cross-entropy for joint training, which jointly
estimates the encoder and classification parameters,
and batch-hard Triplet, SupCon and SoftTriple for
supervised representation learning. These three
criteria adapt
the encoder together with a learned projection
using their respective supervised learning
objectives. The projection produces
128-dimensional representations, which are
normalised to unit length. After adaptation,
the encoder and projection parameters are
kept fixed, and an LR classifier is estimated
using these representations.

The same source-domain partitions are used
for all applicable training and validation
procedures. Segmentation results are reported
for one fitted model per target domain. For the adapted strategies, role
classification results are summarised over five
training runs with different random initialisations.
Model selection follows the domain-averaged
validation criterion defined in
Eq.~\eqref{chap3:eq:source_model_selection}.
The training configurations are detailed in
\ref{chap3:app:model_settings}.

Role classifiers are trained and first
evaluated on reference factual units,
independently of segmentation errors.
In the complete framework, a single fitted
segmenter is retained for each target domain
and its extracted units are used by the
corresponding role classifiers. Thus, the
variation between classification runs is
assessed for fixed segmentations.

\subsection{Evaluation criteria}
\label{chap3:subsec:metrics}

Performance is evaluated separately for each
of the six held-out domains. Overall performance
is reported as the unweighted mean of the
domain-specific scores. For factual-unit extraction, a predicted interval
$\widehat I$ and a reference interval $I$ are compared using
their intersection over union (IoU), defined as
\[
\operatorname{IoU}(\widehat I,I)
=\frac{|\widehat I\cap I|}{|\widehat I\cup I|},
\]
where $|\cdot|$ denotes the number of characters.
This measure quantifies the proportion of
the combined interval covered by both $\widehat I$ and $I$.
Before test evaluation, leading and trailing
whitespace, punctuation, symbols, separators and
control characters are removed from predicted
and reference intervals, without modifying their
interior content. Intervals that become empty
are excluded.

Within each narrative, predicted and reference
intervals are matched one-to-one. A pair is
eligible when $\operatorname{IoU}(\widehat I,I)\geq0.5$.
Let $E(x)$ denote the set of eligible pairs.
A matching $\Gamma\subseteq E(x)$ is a set of
pairs in which each predicted and reference
interval appears at most once.
Let $\mathfrak M(x)$ denote the set of all such
matchings. The selected matching satisfies
\begin{equation}
\Gamma^\star(x)\in
\underset{\Gamma\in\mathfrak M(x)}{\arg\max}
|\Gamma|.
\label{chap3:eq:interval_matching}
\end{equation}

Since reference intervals are non-overlapping, as are
predicted intervals, an interval can have
$\operatorname{IoU}\geq0.5$ with two intervals of the
other set only in the boundary case of exact equality,
which a deterministic rule resolves. The matching is
therefore essentially unique. Neither total IoU nor
role agreement is used as a selection criterion. For a held-out domain $q$, let $H_q$ be the number
of matched pairs, $\widehat G_q$ the number of predicted
units, and $G_q$ the number of reference units
included in the evaluation. The factual-unit F1 is
\begin{equation}
F_{1,q}^{\mathrm{unit}}
=
\frac{2H_q}{\widehat G_q+G_q}.
\label{chap3:eq:unit_f1}
\end{equation}

Matched pairs are counted as true positives,
while unmatched predicted and reference intervals
are counted as false positives and false negatives,
respectively. In the appendix, we also report exact-span
F1, which requires
both interval boundaries to coincide after
applying the test evaluation's edge normalisation
to predicted and reference intervals.
Results for alternative IoU thresholds are
reported in
Table~\ref{chap3:tab:iou_threshold_sensitivity}.

Although factual-unit F1 measures overall recovery,
it does not distinguish the types of segmentation
error. We therefore consider the cross-role merge
rate (CRM), which measures how often an extracted
interval combines distinct facts with different
accident-process roles.

Let $\mathcal X_q$ denote the narratives in the
held-out domain $q$, and $\widehat{\mathcal U}(x)$ the
intervals predicted for narrative $x$ by the evaluated
method, after the same edge normalisation. For
$\widehat I\in\widehat{\mathcal U}(x)$, define
$\mathbf 1_{\mathrm{CRM}}(\widehat I,x)=1$ if there exist
two distinct reference units with different roles,
each having at least half of its length covered
by $\widehat I$, and $\mathbf 1_{\mathrm{CRM}}(\widehat I,x)=0$
otherwise. The coverage of a reference interval $I$ by
$\widehat I$ is measured by $|\widehat I\cap I|/|I|$.
The CRM is
\begin{equation}
\mathrm{CRM}_q
=
\frac{
\displaystyle
\sum_{x\in\mathcal X_q}
\sum_{\widehat I\in\widehat{\mathcal U}(x)}
\mathbf 1_{\mathrm{CRM}}(\widehat I,x)
}{
\displaystyle
\sum_{x\in\mathcal X_q}
|\widehat{\mathcal U}(x)|
}.
\label{chap3:eq:cross_role_merge}
\end{equation}

The CRM therefore measures the proportion of
predicted intervals that substantially overlap
at least two reference units with different
roles. It complements factual-unit F1 by
characterising cross-role merging errors.

For role classification, performance is first
measured on reference factual units, independently
of segmentation errors. We use balanced accuracy
(BA) to account for differences in role frequencies.
For domain $q$, it is defined as
\begin{equation}
\mathrm{BA}_q
=
\frac{1}{|\mathcal R|}
\sum_{r\in\mathcal R}
\frac{\mathrm{TP}_{q,r}}
{\mathrm{TP}_{q,r}+\mathrm{FN}_{q,r}},
\label{chap3:eq:role_ba}
\end{equation}
where $\mathrm{TP}_{q,r}$ and
$\mathrm{FN}_{q,r}$ are the numbers of correctly
and incorrectly classified reference units
belonging to role $r$, respectively.

Alongside mean BA, we report worst-domain BA,
defined as the minimum BA across the six held-out
domains. With repeated training, mean BA is
reported with its standard deviation, while
worst-domain BA is computed for each repetition
and then averaged across repetitions.
Class-specific recall and confusion matrices
provide additional diagnostics.

Finally, the complete framework is evaluated
using automatically extracted and classified
factual units. The one-to-one interval matching
defined above is retained, independently of
predicted roles. A matched pair is considered
correct only when the predicted role also agrees
with the reference annotation.

Let $H_q^{\mathrm{role}}$ denote the number of
matched pairs with correctly assigned roles
under this matching. The role-aware F1 is
\begin{equation}
F_{1,q}^{\mathrm{role}}
=
\frac{2H_q^{\mathrm{role}}}{\widehat G_q+G_q}.
\label{chap3:eq:role_span_f1}
\end{equation}

This criterion requires both a sufficiently
accurate interval and the correct role label.
It therefore evaluates the final structured
representation, whereas balanced accuracy
measures role classification independently
of segmentation. For $H_q>0$, define
$\pi_q=H_q^{\mathrm{role}}/H_q$, the proportion
of matched units receiving the correct role.
Because both F1 measures have the same
denominator, they satisfy the exact identity
\begin{equation}
F_{1,q}^{\mathrm{role}}
=F_{1,q}^{\mathrm{unit}}\,\pi_q.
\label{chap3:eq:two_stage_f1_identity}
\end{equation}

Thus, the end-to-end score separates factual-unit
recovery from role correctness among recovered units.
For comparisons between segmentation methods,
confidence intervals for differences in factual-unit
F1 are obtained by paired bootstrap resampling
of test narratives within each domain. Both methods
are evaluated on the same resampled narratives,
conditional on the fitted segmentation models.

\section{Results and discussion}
\label{chap3:sec:results}

We report factual-unit extraction, role classification
on reference units, and end-to-end recovery of
role-labelled units. Overall scores are unweighted
means across the six held-out domains.

\subsection{Factual-unit extraction}
\label{chap3:subsec:results_extraction}

Table~\ref{chap3:tab:results_segmentation} reports
factual-unit F1, the cross-role merge rate (CRM) and
the ratio of predicted to reference unit counts,
$\rho_q=\widehat G_q/G_q$, averaged over domains as
$\overline{\rho}=\frac{1}{6}\sum_{q=1}^{6}\rho_q$.

\begin{table}[!htbp]
\centering
\caption{Factual-unit extraction across the six held-out
domains. Scores are unweighted domain means, with
unit F1 evaluated at $\operatorname{IoU}\geq0.5$. Best values are in bold.}
\label{chap3:tab:results_segmentation}
\small
\begin{tabular*}{\columnwidth}{@{}l@{\extracolsep{\fill}}ccc@{}}
\toprule
Method & Unit F1 $\uparrow$ & CRM (\%) $\downarrow$ & $\overline{\rho}$ \\
\midrule
PySBD                  & 0.602 & 33.39 & 0.559 \\
SaT, fixed threshold   & 0.615 & 32.54 & 0.570 \\
SaT, source-calibrated & 0.632 & 19.43 & 0.900 \\
SaT, LoRA-adapted      & \textbf{0.847} & \textbf{4.56} & 1.039 \\
\bottomrule
\end{tabular*}
\end{table}

Adapting SaT to factual-unit boundaries raises unit F1
from 0.632 to 0.847 and reduces the CRM from 19.43\% to
4.56\%, compared with source-calibrated SaT. Threshold
calibration alone yields a much smaller gain in unit F1
(from 0.615 to 0.632). Learning task-specific boundary
predictions is therefore the main source of improvement,
both for recovering units and for separating facts with
different roles.

Sentence-level segmenters produce too few units: PySBD
and fixed-threshold SaT yield about 0.56 predicted units
per reference unit, consistent with several facts being
grouped within each sentence. Calibration raises this
ratio to 0.900 and adaptation to 1.039. The count ratio
indicates the direction of discrepancies but does not, by
itself, establish boundary correctness.

\begin{figure}[!ht]
\centering
\includegraphics[width=\columnwidth]
{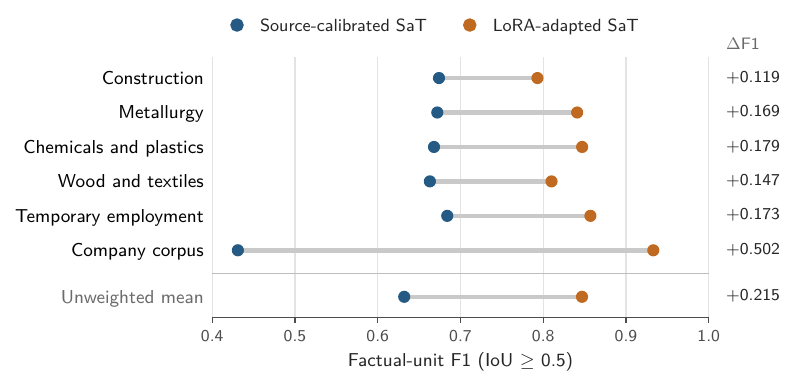}
\caption{Factual-unit F1 ($\operatorname{IoU}\geq0.5$)
by held-out domain for source-calibrated and
LoRA-adapted SaT. The right column reports the
F1 improvement; 95\% bootstrap intervals are
given in Table~\ref{chap3:tab:domain_bootstrap_gain}.}
\label{chap3:fig:domain_gains}
\end{figure}

The gain from adaptation is observed in every held-out
domain (Figure~\ref{chap3:fig:domain_gains}). It ranges
from 0.119 to 0.179 across the five sector-specific
domains (mean 0.157) and reaches 0.502 on the Company
corpus, for an overall mean of 0.215. Since adaptation
and model selection use source-domain data only, these
consistent gains indicate that the learned factual-unit
boundaries transfer to unseen reporting domains. They
persist under stricter IoU thresholds and exact-span
matching (Table~\ref{chap3:tab:iou_threshold_sensitivity}),
and the paired bootstrap intervals of all six domain-level
gains exclude zero (Table~\ref{chap3:tab:domain_bootstrap_gain}).

The Company corpus shows why the CRM must be read
together with unit F1 and the count ratio.
Source-calibrated SaT produces 40,471 intervals for
18,900 reference units ($\rho_q=2.14$) with a CRM of only
3.34\%: it over-segments rather than merges. After
adaptation, the number of intervals falls to 19,507
($\rho_q=1.03$) and unit F1 increases accordingly.
A low CRM alone is therefore not sufficient to assess
factual-unit recovery.

\subsection{Accident-process role classification}
\label{chap3:subsec:results_classification}

Table~\ref{chap3:tab:results_roles} reports role
classification on reference factual units, independently
of segmentation errors, using balanced accuracy (BA) and
macro-F1, the mean of the class-specific F1 scores.

\begin{table}[!ht]
\centering
\caption{Role classification on reference factual
units. Mean BA and macro-F1 are unweighted domain
means; worst BA is the minimum domain-level BA.
For adapted strategies, $\pm$ denotes the standard
deviation over five training runs. Best values are in bold.}
\label{chap3:tab:results_roles}
\small
\begin{tabular*}{\columnwidth}{@{}l@{\extracolsep{\fill}}ccc@{}}
\toprule
Method & Mean BA & Worst BA & Macro-F1 \\
\midrule
TF--IDF + LR & 0.795 & 0.780 & 0.784 \\
Fixed Qwen3 + LR & 0.789 & 0.773 & 0.767 \\
Cross-entropy & $0.842 \pm 0.002$ & 0.824 & \textbf{0.837} \\
Batch-hard Triplet + LR & $0.810 \pm 0.032$ & 0.794 & 0.802 \\
SupCon + LR & $0.835 \pm 0.003$ & 0.817 & 0.820 \\
SoftTriple + LR & $\mathbf{0.845} \pm 0.003$ & \textbf{0.827} & 0.835 \\
\bottomrule
\end{tabular*}
\end{table}

Supervised adaptation of the encoder is the decisive
factor. Fixed Qwen3 representations do not outperform the
lexical TF--IDF baseline (BA 0.789 versus 0.795), whereas
all adapted strategies exceed both. Cross-entropy, SupCon
and SoftTriple reach mean BA of 0.842, 0.835 and 0.845,
respectively, while batch-hard Triplet reaches 0.810 with
greater variation across training runs.

Among the strongest strategies, differences are small.
SoftTriple obtains the highest mean and worst-domain BA,
whereas cross-entropy obtains a slightly higher macro-F1
(0.837 versus 0.835); neither leads on all criteria.
Chemicals and plastics is the most difficult target domain
for all methods except batch-hard Triplet. Role-specific
confusion patterns are reported in
\ref{chap3:app:role_diagnostics}.

\subsection{End-to-end factual-unit and role recovery}
\label{chap3:subsec:results_end_to_end}

The complete framework assigns roles to automatically
extracted units. A predicted unit is correct only when it
is matched to a reference interval and receives the
correct role. For each held-out domain, the same fitted
segmenter is used across the five training runs of each
role classifier.

\begin{table}[!ht]
\centering
\caption{End-to-end role-aware F1, averaged across the
six held-out domains. The upper panel varies the
segmenter for a fixed role classifier; the lower panel
varies the role classifier for a fixed segmenter.
$\pm$ denotes the standard deviation over five training
runs of the role classifier. Best values are in bold.}
\label{chap3:tab:results_e2e_ablations}
\small
\begin{tabular}{@{}lc@{}}
\toprule
Configuration & Role-aware F1 \\
\midrule
\multicolumn{2}{@{}l}{\textit{Role classifier: SoftTriple + LR}} \\
\addlinespace[2pt]
SaT, source-calibrated & $0.515 \pm 0.000$ \\
SaT, LoRA-adapted & $\mathbf{0.703} \pm 0.001$ \\
\midrule
\multicolumn{2}{@{}l}{\textit{Segmentation model: LoRA-adapted SaT}} \\
\addlinespace[2pt]
Cross-entropy & $\mathbf{0.703} \pm 0.002$ \\
SupCon + LR & $0.692 \pm 0.003$ \\
SoftTriple + LR & $\mathbf{0.703} \pm 0.001$ \\
\bottomrule
\end{tabular}
\end{table}

Table~\ref{chap3:tab:results_e2e_ablations} isolates the
contribution of each stage. With the same role classifier,
replacing source-calibrated SaT by the adapted segmenter
raises role-aware F1 from 0.515 to 0.703, a gain of 0.188.
With the same segmenter, the role classifiers differ by at
most 0.011. Boundary estimation therefore has a much larger
effect on the final representation than the choice among
the evaluated classification strategies: when distinct
facts are merged, their separate roles cannot be recovered
by the subsequent classifier.

Table~\ref{chap3:tab:results_domains} details the results
by held-out domain, together with the conditional role
correctness $\pi_q$ defined in
Eq.~\eqref{chap3:eq:two_stage_f1_identity}.

\begin{table}[!ht]
\centering
\caption{Domain-level unit F1, role-aware F1 and
conditional role correctness $\pi_q$ for LoRA-adapted SaT
with SoftTriple + LR ($\operatorname{IoU}\geq0.5$).
Highest values in each column are shown in bold.}
\label{chap3:tab:results_domains}
\small
\begin{tabular*}{\columnwidth}{@{}l@{\extracolsep{\fill}}ccc@{}}
\toprule
Target domain & Unit F1 & Role-aware F1 & $\pi_q$ \\
\midrule
Construction & 0.793 & $0.663 \pm 0.001$ & 0.836 \\
Metallurgy & 0.841 & $0.693 \pm 0.003$ & 0.824 \\
Chemicals and plastics & 0.847 & $0.684 \pm 0.003$ & 0.808 \\
Wood and textiles & 0.810 & $0.650 \pm 0.001$ & 0.802 \\
Temporary employment & 0.857 & $0.709 \pm 0.005$ & 0.827 \\
Company corpus & \textbf{0.933} & $\mathbf{0.818} \pm 0.008$ & \textbf{0.877} \\
\midrule
Unweighted mean & 0.847 & $0.703 \pm 0.001$ & --- \\
\bottomrule
\end{tabular*}
\end{table}

Conditional role correctness ranges from 0.802 to 0.877,
so role errors remain among correctly delimited units.
The decomposition makes these two sources of error
visible: Wood and textiles has a higher unit F1 than
Construction (0.810 versus 0.793) but a lower role-aware F1
(0.650 versus 0.663), because of its lower $\pi_q$. Unit F1
alone therefore does not characterise the quality of the
recovered role-labelled units.

\begin{figure}[!htbp]
\centering
\includegraphics[width=\columnwidth]
{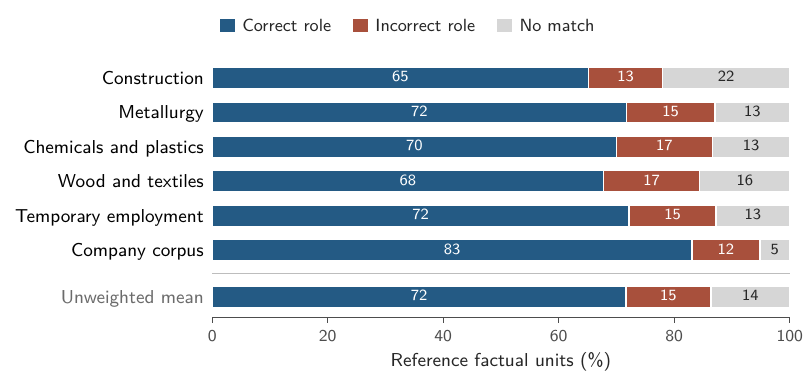}
\caption{Outcome of reference factual units by held-out domain for
LoRA-adapted SaT with SoftTriple + LR: matched with the correct role,
matched with an incorrect role, or not matched (\%).}
\label{chap3:fig:error_decomposition}
\end{figure}

Figure~\ref{chap3:fig:error_decomposition} gives the
outcome of each reference unit. Across domains, about
72\% are recovered with the correct role, 15\% are
matched with an incorrect role and 14\% are not matched,
subject to rounding. The two error types are of comparable
magnitude, which supports evaluating boundary recovery and
role assignment jointly. Since these proportions are
relative to reference units, they do not account for
unmatched predicted intervals, which role-aware F1 does.
Further confusion analyses, including results on matched
extracted units, are reported in
\ref{chap3:app:role_diagnostics}.

\FloatBarrier

\subsection{Limitations and future directions}
\label{chap3:subsec:limitations}

The framework classifies factual units individually,
without explicitly using their surrounding narrative
context; incorporating this context may help distinguish
ambiguous roles. Comparing the two-stage formulation with
joint prediction of boundaries and roles would clarify
their respective strengths. Finally, the evaluation is
restricted to French narratives; extending it to other
languages and reporting systems would further test the
generality of the framework. These directions can be
investigated within the same cross-domain evaluation
setting.

\section{Conclusion}
\label{chap3:sec:conclusion}

This study proposed a two-stage framework that represents
occupational accident narratives as individual reported facts,
each assigned an accident-process role. The framework was
evaluated under multi-source domain generalisation, with all
estimation and model selection performed on source domains only.

Adapting a pretrained sentence-segmentation model to factual-unit
boundaries increased mean factual-unit F1 from 0.632 to 0.847,
with gains in each of the six held-out domains. Supervised
adaptation of the encoder improved role classification, with
cross-entropy and SoftTriple achieving similar performance.
End to end, boundary adaptation raised role-aware F1 from 0.515
to 0.703, whereas the evaluated role classifiers differed by at
most 0.011. Recovering factual units at the appropriate
granularity is therefore the main determinant of the final
representation: facts merged into a single interval cannot
receive separate roles.

The decomposition of role-aware F1 into factual-unit recovery and
conditional role correctness further separates these two sources
of error. By representing reported facts individually, the
framework provides a basis for analysing recurrent accident
circumstances relevant to occupational accident prevention.

\clearpage
\bibliographystyle{elsarticle-harv}
\bibliography{references}
\appendix
\section{Supplementary methods and results}
\label{app:chapter3_factual_units}

\subsection{Reference annotations}
\label{chap3:app:data_splits}

The reference annotations were produced by two experts in
occupational accident analysis. A stratified sample of
approximately 20\% of the narratives from each domain was
annotated independently by both experts: 973 narratives from
Construction, 548 from Metallurgy, 147 from Chemicals and
plastics, 339 from Wood and textiles, 460 from Temporary
employment and 1,118 from the Company corpus, for a total of
3,585 narratives. Each expert identified factual-unit boundaries
and assigned one of the five accident-process roles to each unit.

Agreement was measured on these independent annotations, before
disagreements were resolved. Boundary agreement uses the
one-to-one interval matching of Section~\ref{chap3:subsec:metrics}
with $\operatorname{IoU}\geq0.5$; role agreement is evaluated on
matched units with Cohen's kappa. The two experts identified
29,497 and 29,703 factual units, of which 26,640 were matched,
giving an inter-annotator factual-unit F1 of 0.900. Among matched
units, 92.3\% received the same role (Cohen's kappa 0.900), and
the role-aware F1 was 0.831. Disagreements were then discussed to
refine the annotation guidelines and establish the final
annotations of the shared sample.

The remaining 14,342 narratives were divided between the two
experts, each narrative being annotated by one of them following
the refined guidelines. Ambiguous boundaries or roles were
discussed jointly before finalisation. The final reference corpus
comprises 17,927 narratives and 147,487 factual units.

\FloatBarrier

\subsection{Corpus characteristics}
\label{chap3:app:corpus_characteristics}

Table~\ref{chap3:tab:corpus_lengths} reports narrative and
factual-unit lengths by domain. Narrative length is measured on
the full text and unit length on the annotated span. The Company
corpus contains much shorter narratives and fewer factual units
per narrative than the five sector-specific domains.

\begin{table}[!ht]
\centering
\caption{Narrative and factual-unit lengths by domain, in
characters (median [Q1, Q3]), and mean number of factual units
per narrative.}
\label{chap3:tab:corpus_lengths}
\small
\begin{tabular*}{\columnwidth}{@{}l@{\extracolsep{\fill}}ccc@{}}
\toprule
Domain & Narrative length & Unit length & Units/narrative \\
\midrule
Construction           & 593 [363, 912] & 54 [36, 84] & 10.18 \\
Metallurgy             & 625 [371, 946] & 56 [37, 84] & 10.37 \\
Chemicals and plastics & 701 [466, 947] & 57 [38, 85] & 11.13 \\
Wood and textiles      & 662 [428, 953] & 56 [36, 84] & 10.92 \\
Temporary employment   & 624 [370, 929] & 56 [37, 84] & 10.41 \\
Company corpus         & 129 [93, 170]  & 36 [24, 49] & 3.38 \\
\bottomrule
\end{tabular*}
\end{table}

\FloatBarrier

\subsection{SoftTriple centre normalisation and regularisation}
\label{app:softtriple_details}

The projected representations $z_i\in\mathbb R^{d_p}$ have unit
norm (Section~\ref{chap3:subsec:role_classification}). Each
SoftTriple centre is normalised in the same way,
$\widetilde\mu_{r\ell}=\mu_{r\ell}/\|\mu_{r\ell}\|_2$, so that
$s_{ir\ell}=\cos(z_i,\mu_{r\ell})=z_i^\top\widetilde\mu_{r\ell}$.

The regularisation term penalises distances between centres of
the same role:
\begin{equation}
\mathcal R_{\mathrm{merge}}(\mu)
=\frac{1}{|\mathcal R|\,Q(Q-1)}
\sum_{r\in\mathcal R}
\sum_{1\leq\ell<m\leq Q}
\bigl\|\widetilde\mu_{r\ell}-\widetilde\mu_{rm}\bigr\|_2.
\label{app:eq:softtriple_merge}
\end{equation}
Minimising this term draws centres of the same role together, so
that redundant centres can merge. Weight decay is applied
separately during optimisation.

\FloatBarrier

\subsection{Training settings}
\label{chap3:app:model_settings}

Tables~\ref{chap3:tab:sat_training_settings}
and~\ref{chap3:tab:role_training_settings} report the settings
used for factual-unit segmentation and role classification.

\begin{table}[!ht]
\centering
\caption{Settings for factual-unit segmentation.}
\label{chap3:tab:sat_training_settings}
\small
\begin{tabular*}{\columnwidth}{@{}l@{\extracolsep{\fill}}l@{}}
\toprule
Setting & Value \\
\midrule
Pretrained model & SaT, \texttt{sat-12l-sm} \\
Input block size & 256 tokens \\
\addlinespace[3pt]
LoRA rank $k$, scaling $\alpha$ & $k=16$, $\alpha=32$ \\
Adapted layers & attention and intermediate layers \\
\addlinespace[3pt]
Training & up to 30 epochs, batch size 8 \\
Learning rate & $3\times10^{-4}$, warm-up ratio 0.1 \\
Weight decay & 0 \\
\addlinespace[3pt]
Fixed threshold (pretrained SaT) & 0.25 \\
Calibrated threshold $\tau$ & selected among $\{0, 1/30, \ldots, 1\}$ \\
\bottomrule
\end{tabular*}
\end{table}

\begin{table}[!ht]
\centering
\caption{Settings for accident-process role classification.
Symbols refer to Section~\ref{chap3:subsec:role_classification}.}
\label{chap3:tab:role_training_settings}
\small
\begin{tabular*}{\columnwidth}{@{}l@{\extracolsep{\fill}}l@{}}
\toprule
Setting & Value \\
\midrule
\multicolumn{2}{@{}l}{\textit{Shared settings}} \\
\addlinespace[2pt]
Encoder & \texttt{Qwen/Qwen3-Embedding-0.6B} ($d=1{,}024$) \\
Input length & at most 256 tokens \\
Encoder adaptation & full fine-tuning, 3 epochs, AdamW \\
Learning rates & encoder $2\times10^{-5}$; head or projection $10^{-3}$ \\
Weight decay & 0.01 \\
Projection $\Pi_\psi$ & $d_p=128$, $\ell_2$-normalised \\
Logistic regression & balanced class weights, at most 1,000 iterations; \\
 & inverse regularisation strength in $\{0.01, 0.1, 1, 10\}$ \\
\midrule
\multicolumn{2}{@{}l}{\textit{Method-specific settings}} \\
\addlinespace[2pt]
TF--IDF & word 1--2-grams, min.\ document frequency 2, \\
 & at most 50,000 features, sublinear term frequency \\
Fixed Qwen3 & representations standardised before LR \\
Cross-entropy & batch size 16, inverse-frequency weights $w_r$, dropout 0.1 \\
Batch-hard Triplet & balanced batches of 10 units \\
SupCon & balanced batches of 10 units, $T_c=0.07$ \\
SoftTriple & batch size 16, $Q=5$, $\gamma=0.1$, $\lambda=10$, $\delta=0.01$, $\zeta=0.2$ \\
\bottomrule
\end{tabular*}
\end{table}

\FloatBarrier

\subsection{Additional segmentation results}
\label{chap3:app:additional_segmentation}

To quantify the uncertainty of the segmentation gains, we used a
paired bootstrap within each held-out domain. Test narratives were
resampled with replacement 5,000 times; for each sample, the
factual-unit F1 of source-calibrated and LoRA-adapted SaT was
recomputed on the same narratives and their difference recorded.
The 2.5th and 97.5th percentiles of these differences give the
95\% intervals in Table~\ref{chap3:tab:domain_bootstrap_gain}.
These intervals reflect sampling variability conditional on the
fitted models.

\begin{table}[!ht]
\centering
\caption{Gain in factual-unit F1 ($\operatorname{IoU}\geq0.5$) of
LoRA-adapted over source-calibrated SaT, with 95\% paired
bootstrap intervals.}
\label{chap3:tab:domain_bootstrap_gain}
\small
\begin{tabular*}{\columnwidth}{@{}l@{\extracolsep{\fill}}cc@{}}
\toprule
Held-out domain & F1 gain & 95\% interval \\
\midrule
Construction           & $+0.119$ & $[0.112,\,0.126]$ \\
Metallurgy             & $+0.169$ & $[0.161,\,0.176]$ \\
Chemicals and plastics & $+0.179$ & $[0.162,\,0.196]$ \\
Wood and textiles      & $+0.147$ & $[0.136,\,0.157]$ \\
Temporary employment   & $+0.173$ & $[0.165,\,0.181]$ \\
Company corpus         & $+0.502$ & $[0.493,\,0.511]$ \\
\bottomrule
\end{tabular*}
\end{table}

Table~\ref{chap3:tab:iou_threshold_sensitivity} evaluates the
same predictions under other IoU thresholds and exact-span
matching, with models and thresholds unchanged. LoRA-adapted SaT
remains better under every criterion, and the gap widens as the
matching becomes stricter.

\begin{table}[!ht]
\centering
\caption{Mean factual-unit F1 across the six held-out domains
under different matching criteria, with the same edge
normalisation throughout.}
\label{chap3:tab:iou_threshold_sensitivity}
\small
\begin{tabular}{@{}lcc@{}}
\toprule
Matching criterion & Source-calibrated SaT & LoRA-adapted SaT \\
\midrule
$\operatorname{IoU}\geq 0.3$ & 0.720 & 0.884 \\
$\operatorname{IoU}\geq 0.5$ & 0.632 & 0.847 \\
$\operatorname{IoU}\geq 0.7$ & 0.439 & 0.709 \\
Exact span                   & 0.371 & 0.663 \\
\bottomrule
\end{tabular}
\end{table}

\FloatBarrier

\subsection{Role-classification diagnostics}
\label{chap3:app:role_diagnostics}

\begin{figure*}[!ht]
\centering
\begin{subfigure}[t]{0.46\textwidth}
\centering
\includegraphics[width=\linewidth]
{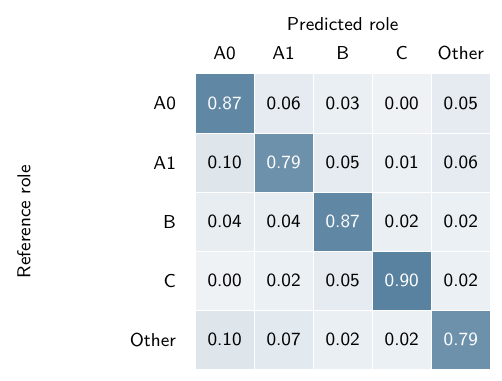}
\caption{Reference factual units.}
\end{subfigure}\hfill
\begin{subfigure}[t]{0.46\textwidth}
\centering
\includegraphics[width=\linewidth]
{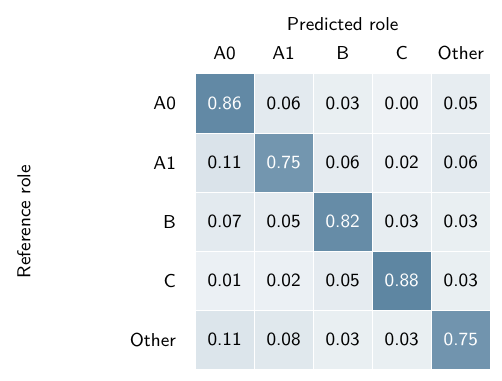}
\caption{Matched extracted units.}
\end{subfigure}
\caption{SoftTriple role-confusion matrices on (a) reference
factual units and (b) matched units extracted by LoRA-adapted SaT.
Row-normalised proportions are averaged over six held-out domains
and five training runs.}
\label{chap3:fig:role_confusion}
\end{figure*}

Figure~\ref{chap3:fig:role_confusion} compares SoftTriple
confusion matrices on reference factual units and on extracted
units matched to a reference unit; unmatched units are excluded
from the second matrix. For each target domain and training run,
the matrix is normalised by reference role, then averaged over
the six domains and five runs.

\FloatBarrier

\end{document}